\documentclass{article}
\usepackage[T1]{fontenc}
\usepackage{iclr2027_conference,times}

\usepackage{amsmath,amsfonts,bm}

\def\eqref#1{\ref{#1}}

\def\1{\bm{1}}

\DeclareMathAlphabet{\mathsfit}{\encodingdefault}{\sfdefault}{m}{sl}
\SetMathAlphabet{\mathsfit}{bold}{\encodingdefault}{\sfdefault}{bx}{n}

\usepackage{graphicx}
\usepackage{booktabs}
\usepackage{pifont}
\usepackage{array}
\usepackage{tabularx}
\usepackage{wasysym}
\usepackage{xcolor}
\usepackage{hyperref}
\usepackage{xurl}
\usepackage{wrapfig}
\usepackage{capt-of}
\usepackage{placeins}
\usepackage{needspace}
\graphicspath{{figures/}}
\newlength{\sidepanelwidth}
\title{Sekai: Incremental Event Memory for \\Generalized Reasoning Video Segmentation}
\author{Zhengtao Liu$^{1}$ \quad Jinrong Xu$^{1}$ \quad Shiu-Hong Kao$^{2}$\\
\bfseries Huixin Sun$^{2}$ \quad Yicong Li$^{1}$\thanks{Corresponding author: \href{mailto:liyicong@ustc.edu.cn}{\texttt{liyicong@ustc.edu.cn}}.}\\
\normalfont $^{1}$University of Science and Technology of China\\
\normalfont $^{2}$National University of Singapore}
\iclrfinalcopy
\makeatletter
\let\sekai@maketitle\@maketitle
\renewcommand{\@maketitle}{\sekai@maketitle\lhead{Preprint}}
\makeatother
\hypersetup{
  hidelinks,
  pdftitle={Sekai: Incremental Event Memory for Generalized Reasoning Video Segmentation},
  pdfauthor={Zhengtao Liu, Jinrong Xu, Shiu-Hong Kao, Huixin Sun, Yicong Li}
}

\begin{document}
\maketitle
\begin{abstract}

Online reasoning video segmentation (Online RVS) requires a model to segment query-referred objects as a video streams in, using only the frames observed so far. Existing formulations largely assume a fixed target set throughout the video, overlooking a fundamental property of streaming reasoning: as new events occur, the objects that satisfy a query may change over time. To address this, we propose a new task, \textbf{Generalized Reasoning Video Segmentation (GRVS)}, where the target set depends on query-relevant events accumulated over the observed video history. As new evidence accumulates, the target set can evolve from empty to one or multiple objects, requiring a model to continuously maintain target validity while localizing the valid objects currently in view. 
To evaluate GRVS, we build \textbf{Isekai}, a diagnostic benchmark of 60 videos and 197 queries across 11 temporal query families, with 12{,}569 scoring points whose target sets are causally annotated over time. We further propose \textbf{Sekai}, an online framework with two complementary components: an \textbf{Event Memory Reasoner (EMR)} that maintains query-relevant events and evolving target states in structured memory, and a \textbf{Memory-Grounded Decoder (MGD)} that identifies valid targets from this memory and localizes them in the current frame. Experiments show that Sekai significantly outperforms existing online RVS methods, achieving 50.33 J\&F and improving over the strongest baseline by 3.2 points. Code, example annotations, and evaluation scripts are available at \url{https://github.com/zenastangbi/Sekai}.
\end{abstract}
\suppressfloats[t]
\begin{figure}[t]
\centering
\includegraphics[width=132.5mm,trim=3bp 3bp 3bp 2.5bp,clip]{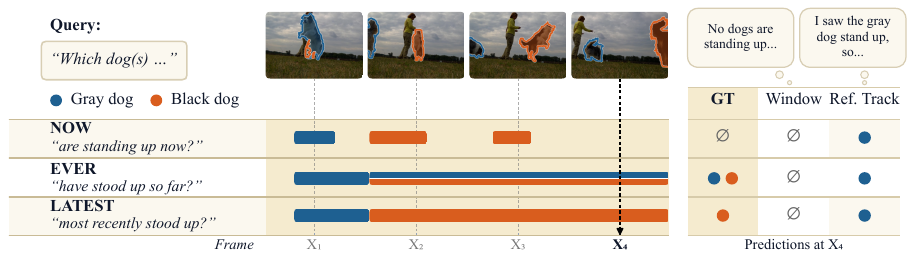}
\caption{\textbf{Generalized Reasoning Video Segmentation (GRVS).} Target identities dynamically evolve under causal temporal contexts, leading to varying targets that standard sliding-window or tracking-based baselines fail to capture.}
\label{fig:task}
\end{figure}

\section{Introduction}
\label{sec:introduction}

Online Reasoning Video Segmentation (Online RVS) brings language-guided segmentation to streaming video, where the system can only access frames observed so far and must produce segmentation results on the fly. Unlike offline settings with the complete video available, the model in online RVS cannot access future context or defer its decision to the end of the sequence. Online segmentation is thus essential for real-world applications that must continuously adapt to an unfolding visual stream.

However, the online setting exposes a fundamental yet largely overlooked challenge: the segmentation target can evolve as the video unfolds. Consider the example in Figure~\ref{fig:task}, where the query asks, ``Which dogs have stood up?''. Before either dog stands up, there is no valid target; after the first dog stands up, only that dog should be segmented; once the second dog also stands up, both dogs become valid targets. More generally, the target set can grow, shrink, or switch between instances as new evidence arrives. Two common online paradigms struggle with such evolving targets. Reference-tracking methods ground the referred object once and then follow it, regardless of whether other instances later satisfy the query, and thus miss the second dog. Window-based methods ground the query only within the current observation window; once the first dog's standing-up event leaves the window, they output an empty mask, even though the dog remains a valid target.

Yet existing online methods are largely built around a fixed-target formulation and cannot adapt to evolving target sets. Both paradigms above share a common cause: they do not explicitly record which objects have already satisfied the query and what events support that decision. Window-based methods discard evidence once it leaves the current window. Reference-tracking methods use temporal memory mainly to propagate the identified target across frames, through visual features or cached tokens. Such memory captures what the target looks like, but not how the target set changes as new events occur. The same limitation appears at the evaluation level: existing benchmarks typically assign a fixed target set to the entire video, and thus cannot directly test whether a model correctly follows changes in the valid target set over time.

Therefore, we formalize a new task, \textbf{Generalized Reasoning Video Segmentation} (GRVS). In GRVS, the model processes the video as a stream. At each moment, the valid target set is determined by the query-relevant events observed so far, rather than by a one-time reference or the current window alone. As the stream evolves, this set may change in both size and membership. Importantly, target validity is also distinct from instantaneous visibility: an object may remain a valid target even when temporarily occluded or outside the frame, and should be segmented again once it becomes visible. GRVS thus generalizes reasoning video segmentation in three aspects: it operates causally on a stream, allows the target set to vary in size and over time, and separates target validity from visibility. These properties give rise to two fundamental requirements: maintaining the evolving target state over time and localizing the currently visible valid targets at each moment.

To evaluate GRVS, we construct \textbf{Isekai}, a diagnostic benchmark of 60 real videos, 197 queries, and 12,569 scored query-frame instances. Unlike existing benchmarks, Isekai annotates targets causally: the valid target set at each frame depends only on what has been observed up to that frame, and thus changes over time. The queries span 11 temporal query families, ranging from current states, past occurrences, and most recent events to absent events, event counts, and temporal order. Instances are further stratified by how long ago the supporting evidence occurred, which allows us to measure how well a model retains past evidence. 

To meet these two requirements, we introduce \textbf{Sekai}, an online framework whose name summarizes its design: knowledge of query-relevant events is accumulated in increments, one arriving chunk at a time, in a structured text memory, and segmentation is decoded from this knowledge rather than from the current window alone. As each video chunk arrives, the Event Memory Reasoner (EMR) maintains the evolving target state by recording query-relevant events in this memory, keeping, for each object, what it has done and when. The Memory-Grounded Decoder (MGD) then reasons over the memory to decide which objects currently satisfy the query, and localizes those visible in the current frame. Experiments on Isekai show that Sekai consistently outperforms existing online methods, demonstrating its effectiveness in reasoning over dynamically evolving targets.

Our contributions are:
\begin{itemize}

\item We formalize GRVS, a streaming task in which query-relevant events observed so far determine a time-varying target set of variable size, and target validity is decoupled from instantaneous visibility. 

\item We introduce Isekai, a diagnostic benchmark for GRVS spanning 11 temporal query families. To the best of our knowledge, it is the first benchmark to provide causally defined, time-varying target annotations.

\item We propose Sekai, an online segmentation framework for GRVS that records query-relevant events in a structured text memory, and decodes the resulting variable-sized target set into current-frame masks.
\end{itemize}
\section{Related Work}
\label{sec:related_work}

\paragraph{Reasoning Video Segmentation.}
Referring video object segmentation grounds language expressions to video masks, with motion-centric expressions making temporal dynamics an explicit part of the query~\citep{ding2023mevis}. Reasoning variants extend expressions to implicit descriptions that require world knowledge or multi-step inference, typically by decoding multimodal language model representations into masks or tracks~\citep{bai2024videolisa,yan2024visa,zheng2025villa,munasinghe2025videoglamm,gong2025vrshq,yuan2026sa2va}. A second line decouples semantic selection from spatial-temporal propagation, either by prompting propagation from reasoned keyframes~\citep{kao2026cotrvs} or by pairing candidate selection with dedicated tracking stages~\citep{jin2026agentrvos,zhu2026sdam}. Generalized referring segmentation introduced empty- and multi-target outputs for images (GRES~\citep{liu2023gres}, GSVA~\citep{xia2024gsva}), but video formulations still assume static instance identities across the sequence. GRVS instead makes eligibility a function of prefix-confirmed events.

\paragraph{Memory in Video Segmentation.}
 Existing streaming trackers and reasoning frameworks propagate implicit states across time: target queries for selected instances~\citep{wu2023onlinerefer}, visual and mask memories built on SAM2~\citep{cuttano2025samwise,ravi2024sam2}, recurrent language embeddings or token reservoirs~\citep{zhang2026streamingrvos,liu2026orvos}, or re-evaluated keyframe contexts~\citep{kao2026cotrvs}. Long-video models similarly retain visual memory banks or compressed KV caches~(\citealp{he2024malmm,song2024moviechat,di2025rekv,zhang2026hermes,xiao2026mukv}; \mbox{\citealp{shen2025jit}}). However, these approaches rely on continuous visual features or implicit latent tokens, which struggle to preserve symbolic event histories once visual evidence exits the local temporal window. Sekai instead carries an explicit, editable text memory of objects and prefix-confirmed events (Section~\ref{sec:system_record}).

\paragraph{Benchmarks.}
Referring and reasoning VOS benchmarks, including ReVOS~\citep{yan2024visa}, MeViS v2~\citep{ding2025mevis}, and ReasonVOS~\citep{bai2024videolisa}, associate each expression with static instance identities across the video; recent extensions add motion intervals~\citep{deng2025groundmore} or causal referent shifts~\citep{liu2026orvos}, but allow at most one target at a time, and none annotates the events that establish eligibility. Isekai annotates these events directly and derives time-varying, zero-to-many target sets from them (Table~\ref{tab:benchmark_positioning}).

\begin{table}[!t]
\centering
\caption{\textbf{Benchmark annotation.} Comparison of annotations in existing benchmarks and Isekai.}
\label{tab:benchmark_positioning}
\begingroup\scriptsize
\setlength{\tabcolsep}{2.8pt}\renewcommand{\arraystretch}{1.2}
\begin{tabular*}{\linewidth}{@{}l@{\extracolsep{\fill}}ccccccc@{}}
\toprule
\textbf{Benchmark} &
\shortstack{Causal\\labels} &
Empty &
Multi &
\shortstack{Dynamic\\membership} &
\shortstack{Evidence\\age} &
\shortstack{Temporal\\operators} &
\shortstack{Temporal\\grounding} \\
\midrule

\mbox{ReasonVOS~\citep{bai2024videolisa}}
& \ding{55} & \ding{55} & \ding{55} & \ding{55}
& \ding{55} & \ding{55} & Object tracks \\

\mbox{ReVOS~\citep{yan2024visa}}
& \ding{55} & \ding{51} & \ding{51} & \ding{55}
& \ding{55} & \ding{55}
& \shortstack{Object tracks} \\

\mbox{MeViS v2~\citep{ding2025mevis}}
& \ding{55} & \ding{51} & \ding{51} & \ding{55}
& \ding{55} & \ding{55}
& \shortstack{Object tracks} \\

\mbox{GroundMoRe~\citep{deng2025groundmore}}
& \ding{55} & \ding{51} & \ding{51} & \ding{51}
& \ding{55} & \ding{55} & Motion intervals \\

\mbox{ORVOSB~\citep{liu2026orvos}}
& \ding{51} & \ding{51} & \ding{55} & \ding{55}
& \ding{55} & \ding{55}
& \shortstack{Validity intervals} \\

\textbf{Isekai}
& \ding{51} & \ding{51} & \ding{51} & \ding{51}
& \ding{51} & \ding{51}
& \shortstack{Event records} \\
\bottomrule
\end{tabular*}

\par\endgroup
\end{table}

\section{Isekai: Diagnostic Benchmark}
\label{sec:isekai}

\subsection{Construction}
\label{sec:isekai_construction}
\paragraph{Sources and scale.}
Isekai builds on videos and instance masks from MeViS v2~\citep{ding2025mevis}, GroundMoRe~\citep{deng2025groundmore}, and DAVIS 2017~\citep{ponttuset2017davis}: 60 videos, 197 fixed queries, and 12,569 scoring points (Table~\ref{tab:isekai_dataset_statistics}). Isekai is evaluation-only; Sekai is neither trained nor tuned on it (Section~\ref{sec:experimental_setup}).

\begin{figure}[!t]
\centering
\includegraphics[width=0.819\linewidth,trim=4bp 0bp 4bp 0bp,clip]{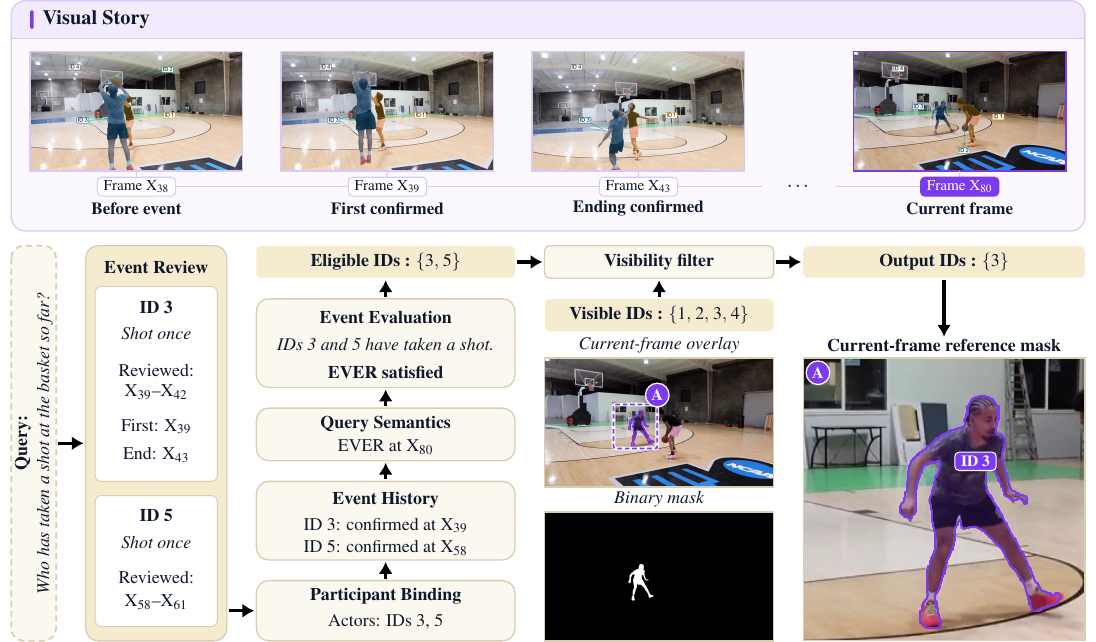}
\caption{\textbf{Isekai annotation example.} A reviewed event is bound to its participants; the \textsc{Ever} predicate selects the eligible IDs, whose intersection with the visible IDs gives the current-frame reference masks.}
\label{fig:isekai_construction}
\end{figure}

\paragraph{Reviewed event timelines.}
Starting from automatically proposed object--event intervals, annotators verify each event's participants and boundaries and mark the first frame at which the prefix confirms it. An event is marked \textsc{ended} only with observed ending evidence, and events whose order cannot be resolved are excluded from scoring. The target set at a frame is the eligible instances that are also visible in the official masks (Figure~\ref{fig:isekai_construction}; Appendix~\ref{app:isekai}).

\paragraph{Annotation agreement.}
A second annotator, without access to the original labels, re-annotated the events in 20 videos. On the 2{,}307 resulting \textsc{Now}/\textsc{Ever}/\textsc{Never} scoring points (Section~\ref{sec:query_families}), the two annotations give identical target sets on 93.0\% (mean Jaccard $0.940$), although 80.6\% of individual events match one-to-one (Appendix~\ref{app:isekai_agreement}). Target sets are thus more stable than the event segmentations that produce them.

\paragraph{Dataset landscape.}
Figure~\ref{fig:isekai_landscape} shows how scoring points distribute over query group, evidence age, and target count; Order pools \textsc{Latest}, \textsc{First}, and \textsc{Before/After}, and Other pools the remaining compound families. Of the 12,569 points, 5,379 have an empty target set and 1,028 have more than one target, so both departures from the single-target setting are common.

\subsection{Query families}
\label{sec:query_families}
Isekai has eleven query families (Table~\ref{tab:families}). The five primary families apply one predicate each to the event history: \textsc{Now} selects the current participants, \textsc{Ever} all participants so far, \textsc{Never} their complement, and \textsc{Latest} and \textsc{First} the most recent and the earliest one.

\noindent The six compound families add counting, ordering, role chains, and duration; ``Who was the first player to take a shot after dribbling the ball themselves?'' must bind the shooter to an earlier dribbling event by the \emph{same} instance. Appendix~\ref{app:isekai_definitions} gives the formal predicates and boundary cases. Results pool the 38 queries outside \textsc{Now}, \textsc{Ever}, \textsc{Latest}, and \textsc{Never} as \emph{Other}.

\WFclear
\subsection{Diagnostic protocol and metrics}
\label{sec:isekai_protocol}
\paragraph{Evidence age and history subset.}
Evidence age is the number of frames from the supporting event's confirmation to the scoring frame, in four strata: in window (confirmed within the current chunk), 1--8, 9--24, and 25+. A scoring point is in the \emph{history subset} if its target set changes when recomputed from only the events confirmed in the current chunk, i.e., its answer depends on evidence outside the window. The subset is defined on \textsc{Now}, \textsc{Ever}, and \textsc{Never}, whose targets follow from the event rules, and contains 4{,}133 of its 8{,}980 points.

\paragraph{Metrics.}
We report region IoU $\mathcal J$ and boundary $\mathcal F$, computed at each scoring point between the union of predicted masks and the union of reference masks; two empty unions score $\mathcal J=\mathcal F=1$, one empty union scores zero. Scores are averaged within each video and then across videos; subgroup results are filtered to their points before averaging. Every frame is scored, so a system that updates its targets only at chunk endpoints is penalized on each frame between a membership change and its next update.

\newsavebox{\isekaistatbox}
\newsavebox{\isekailandscapebox}
\sbox{\isekaistatbox}{%
\begin{minipage}[t]{\dimexpr\sidepanelwidth-\columnsep\relax}
\vspace{0pt}
\centering\small
\captionof{table}{\textbf{Statistics of Isekai.}}
\label{tab:isekai_dataset_statistics}
\scriptsize
\setlength{\tabcolsep}{0pt}
\renewcommand{\arraystretch}{1.08}
\begin{tabular}{@{}>{\centering\arraybackslash}p{0.27450\linewidth}>{\centering\arraybackslash}p{0.172365\linewidth}>{\centering\arraybackslash}p{0.183150\linewidth}>{\centering\arraybackslash}p{0.177835\linewidth}>{\centering\arraybackslash}p{0.192150\linewidth}@{}}
\toprule
Source & Videos & Queries & Frames & Points \\
\midrule
GroundMoRe & 41 & 143 & 2,550 & 9,010 \\
MeViS v2 & 7 & 23 & 450 & 1,448 \\
DAVIS 2017 & 12 & 31 & 823 & 2,111 \\
\midrule
Total & 60 & 197 & 3,823 & 12,569 \\
\bottomrule
\end{tabular}
\end{minipage}}
\sbox{\isekailandscapebox}{%
\begin{minipage}[t]{\sidepanelwidth}
\vspace{0pt}
\centering\small
\includegraphics[width=0.85\linewidth,trim=0.25bp 0.5bp 0.5bp 3bp,clip]{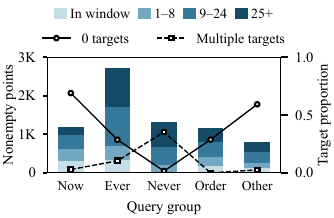}
\captionof{figure}{\textbf{Landscape of Isekai.} Bars count nonempty points by evidence age; the lines show zero-/multiple-target proportions over all points in each group. Appendix~\ref{app:isekai_landscape_timing} gives timing and grouping rules.}
\label{fig:isekai_landscape}
\end{minipage}}
\edef\isekaiwraplines{\the\numexpr\number\dimexpr\ht\isekailandscapebox+\dp\isekailandscapebox-\ht\isekaistatbox-\dp\isekaistatbox+2\intextsep\relax/\number\baselineskip+1\relax}
\Needspace{\dimexpr\ht\isekailandscapebox+\dp\isekailandscapebox+2\intextsep\relax}
\begin{wrapfigure}[\isekaiwraplines]{r}{\sidepanelwidth}
\vspace{\abovecaptionskip}
\usebox{\isekailandscapebox}
\end{wrapfigure}
\noindent\usebox{\isekaistatbox}\par\smallskip

\section{Method}
\label{sec:method}

\paragraph{GRVS Task Definition.}
\label{sec:problem_setup}
A fixed query $q$ is given before the video stream $x_{0:T}$. At time $t\in\{0,\ldots,T\}$ the model sees only the prefix $x_{0:t}$ and cannot revise masks it has already emitted. Let $\mathcal{I}_t$ be the instances observed so far, $s_{i,t}\in\{0,1\}$ indicate whether $i$ satisfies $q$ given $x_{0:t}$, and $v_{i,t}\in\{0,1\}$ whether $i$ is visible at $t$. The target set and the required output are
\WFclear
\begin{equation}
\mathcal{A}_t=\{i\in\mathcal{I}_t : s_{i,t}=1\},\qquad
\mathcal{Y}_t=\{(i,m_{i,t}) : i\in\mathcal{A}_t,\ v_{i,t}=1\},
\label{eq:answer_set}
\end{equation}
where $m_{i,t}\in\{0,1\}^{H_t\times W_t}$ is a nonempty mask of the visible region. Eligibility and visibility are decoupled: an event can flip $s_{i,t}$ while $i$ stays in view, and occlusion can flip $v_{i,t}$ while $i$ stays eligible. Hence $|\mathcal{Y}_t|$ may be zero, one, or many, and an empty output states that no \emph{visible} instance currently qualifies.

Because $q$ refers to events rather than appearance, $s_{i,t}$ depends on history. An event $e=(P_e,[a_e,b_e],\tau_e)$ has participants $P_e\subseteq\mathcal{I}_t$, an interval $[a_e,b_e]$, where $b_e$ is the last active frame, set only once the ending is observed at $b_e+1$ ($b_e=\infty$ if no ending is observed), and a confirmation frame $\tau_e\ge a_e$, the first frame at which the prefix suffices to confirm $e$. The causal timeline $\mathcal{E}_t=\{e:\tau_e\le t\}$ grows monotonically, and every query family is a predicate over it (Appendix~\ref{app:isekai_definitions}).

\begin{figure}[t]
\centering
\includegraphics[width=0.9\linewidth]{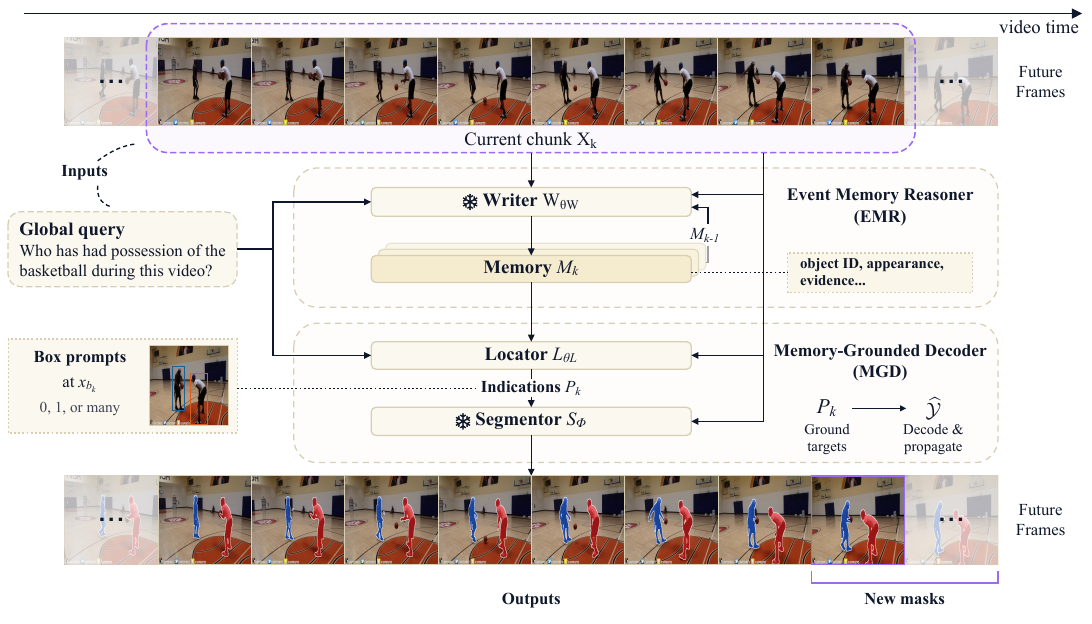}
\caption{\textbf{Sekai.} The EMR maintains event memory; the MGD emits zero, one, or many boxes $P_k$ and decodes and propagates their masks. Snowflakes mark frozen components.}
\label{fig:architecture}
\end{figure}

\paragraph{Framework Overview.}
Sekai splits the two requirements of Equation~\ref{eq:answer_set}, historical eligibility and current localization, between two modules (Figure~\ref{fig:architecture}). The Event Memory Reasoner (EMR) is a frozen writer $W_{\theta_W}$ that maintains a structured text memory $M_k$, a bounded approximation of $\mathcal{E}_t$, across chunks. The Memory-Grounded Decoder (MGD) is a LoRA-adapted locator $L_{\theta_L}$ followed by a frozen segmentor $S_{\Phi}$; it reads $M_k$ against the current chunk and produces the target masks.

After an initialization call on $x_0$, the stream is consumed in chunks of 8 frames, $X_k=x_{a_k:b_k}$ (indexing in Appendix~\ref{app:response_contract}). At call $k$,
\begin{equation}
\begin{aligned}
\text{EMR:}\quad & M_k=W_{\theta_W}(X_k,q,M_{k-1}),\\
\text{MGD:}\quad & P_k=L_{\theta_L}(X_k,q,M_k)=\{p_{k,j}\}_{j=1}^{n_k},\quad
\widehat{\mathcal{Y}}_{b_k:b_{k+1}-1}=S_{\Phi}(x_{b_k:b_{k+1}-1},P_k),
\end{aligned}
\label{eq:joint_update}
\end{equation}
with $M_0=\emptyset$; $P_k$ holds $n_k\ge 0$ boxes on the endpoint frame $x_{b_k}$, and the last propagation interval runs to the end of the video. Writer and locator start from a fresh context at every call, with no KV cache or visual features carried over, so $M_k$ is the only semantic state passed between the modules or across chunks; the segmentor alone keeps a visual propagation state.

\subsection{Event Memory Reasoner}
\label{sec:system_record}
The EMR records event history in an editable text memory so that it survives after the supporting frames have left the window.

\paragraph{Memory structure.}
Each object entry holds a stable local ID, an appearance phrase, its last visible location and frame, and a list of events; each event holds an action label, a state in $\{\textsc{active},\textsc{ended},\textsc{unknown}\}$, and its first and most recent evidence frames. The rendered memory is capped at 128 tokens, so the carried state is $O(1)$ in stream length. For the video in Figure~\ref{fig:task}, the memory at frame $x_4$ reads
\begin{quote}\footnotesize
\begin{verbatim}
{"objects":[
 ["o1","gray dog","left of frame",53,[["stand-up","ended",5,11]]],
 ["o2","black dog","right image boundary",53,
   [["stand-up","ended",15,22],["stand-up","ended",31,37]]]]}
\end{verbatim}
\end{quote}
where 11, 22, and 37 are the frames at which each rearing was first seen to have ended.

\paragraph{Reasoning operations.}
At each call the writer performs three prompt-specified operations. (i) \emph{Identity binding}: keep local IDs stable across calls and never transfer one object's history to another. (ii) \emph{State transition}: an event is \textsc{active} while visibly ongoing, \textsc{ended} only once its ending has been observed, and \textsc{unknown} otherwise; disappearance alone does not end an event, matching the Isekai annotation rule (Appendix~\ref{app:isekai_definitions}). (iii) \emph{Consolidation}: keep supported history and uncertainty, and do not invent events.

\paragraph{Relation to the causal timeline.}
Each field approximates one quantity of an event (owner $\leftrightarrow P_e$, entry presence $\leftrightarrow \tau_e\le t$, \textsc{active}/\textsc{ended} $\leftrightarrow t\in[a_e,b_e]$, first evidence $\leftrightarrow a_e$), so every primary predicate (Eq.~\ref{eq:semantics}, Appendix~\ref{app:isekai_definitions}) has a direct counterpart on $M_k$.

\subsection{Memory-Grounded Decoder}
\label{sec:training}
The MGD turns the EMR's memory into masks on the current frames. We call this memory-grounded decoding: the query is applied to the memory, not to the chunk alone.

\paragraph{Memory readout and grounding.}
The locator reads $q$, $M_k$, and $X_k$ and returns boxes on the endpoint frame $x_{b_k}$ for every visible instance that satisfies $q$. This involves two steps: applying the query's temporal condition to the memory entries, which determines who qualifies, and matching the qualifying entries' appearance phrases and last locations against $x_{b_k}$, which determines where they are. An empty list is a valid output and means no visible instance currently qualifies.

\paragraph{Mask decoding and propagation.}
$S_{\Phi}$ converts the $n_k\ge 0$ boxes into masks on $x_{b_k}$ and propagates them over arriving frames until the next endpoint, where new prompts replace them. Because prompts are recomputed from $M_k$ at every endpoint, a change in the target set, such as a new \textsc{Ever} participant or an ended \textsc{Now} event, takes effect at the next endpoint without revising earlier outputs.

\paragraph{Training.}
We freeze the writer, the vision backbone, and the segmentor, and adapt only the locator with LoRA under response-only cross-entropy on the serialized boxes, $\mathcal{L}=-\tfrac{1}{|y^*|}\sum_r\log p_{\theta_L}(y^*_r\mid X_k,q,M_k,y^*_{<r})$. Training pairs combine writer-generated memories with annotated endpoint boxes. They are supplemented by synthetic memories with substituted event states and query families, which are intended to make the target depend on the memory rather than on the chunk alone. Data and hyperparameters are given in Appendix~\ref{app:locator_config}.

\section{Experiments}
\label{sec:experiments}

\subsection{Experimental Setup}
\label{sec:experimental_setup}
\paragraph{Data and splits.}
We evaluate on Isekai (Section~\ref{sec:isekai}) and, with their original queries and masks, on MeViS v2~\citep{ding2025mevis}, ReasonVOS~\citep{bai2024videolisa}, and ReVOS~\citep{yan2024visa}. MeViS v2 is in-domain for the locator; ReVOS uses its full validation set.

\paragraph{Implementation.}
The EMR's writer and the MGD's locator are both Qwen3-VL-8B-Instruct~\citep{bai2025qwen3vl}; only the locator is LoRA-adapted, on MeViS v2 training data disjoint from Isekai (Appendix~\ref{app:locator_config}). SAM~3.1~\citep{carion2025sam3} is the segmentor. Isekai is scored as in Section~\ref{sec:isekai_protocol}; the other benchmarks use their official metrics on causal predictions.

\begin{table}[!t]
\centering\footnotesize
\caption{\textbf{Online methods on Isekai under identical causal inputs.} Family columns report J\&F and mix empty and nonempty frames in different proportions (Figure~\ref{fig:isekai_landscape}). \emph{Other} pools the families outside Now/Ever/Latest/Never; \emph{Hist.} is the history subset (Section~\ref{sec:isekai_protocol}). }
\label{tab:isekai_online}
\scriptsize
\setlength{\tabcolsep}{2.3pt}
\begin{tabular*}{\linewidth}{@{}l@{\extracolsep{\fill}}rrrrrrrrr@{}}
\toprule
Method & J & F & J\&F & Now & Ever & Latest & Never & Other & Hist. \\
\midrule
OnlineRefer~\citep{wu2023onlinerefer} & 28.08 & 33.38 & 30.73 & 20.92 & 39.47 & 34.75 & 29.10 & 25.93 & 48.87 \\
SAMWISE~\citep{cuttano2025samwise} & 33.52 & 37.07 & 35.30 & 22.30 & 46.94 & 40.17 & 34.07 & 27.52 & 57.93 \\
CoT-RVS~\citep{kao2026cotrvs} & 46.43 & 47.90 & 47.16 & \textbf{59.31} & 39.52 & 25.81 & 18.00 & \textbf{53.63} & 22.15 \\
StreamingRVOS~\citep{zhang2026streamingrvos} & 39.34 & 43.23 & 41.28 & 29.40 & 55.18 & 47.26 & \textbf{38.17} & 33.28 & \textbf{68.71} \\
\midrule
Sekai w/o memory & 44.33 & 46.89 & 45.61 & 34.60 & 59.88 & \textbf{50.73} & 37.51 & 36.43 & 63.93 \\
\textbf{Sekai} & \textbf{49.09} & \textbf{51.58} & \textbf{50.33} & 42.04 & \textbf{64.46} & 44.76 & 27.00 & 48.42 & 65.97 \\
\bottomrule
\end{tabular*}
\end{table}

\paragraph{Baselines and controls.}
We adapt four online methods to the same arrived-frame schedule and audit every input against future frames (Appendix~\ref{app:baselines}): OnlineRefer~\citep{wu2023onlinerefer} propagates a target query, SAMWISE~\citep{cuttano2025samwise} a causal visual window, and StreamingRVOS~\citep{zhang2026streamingrvos} semantic tokens; our CoT-RVS~\citep{kao2026cotrvs} adaptation shares Sekai's Qwen3-VL-8B and SAM~3.1 and rereads a compressed grid of the arrived prefix at each refresh. Two controls bracket Sekai: \emph{Sekai w/o memory} removes the EMR, so the locator sees only $q$ and $X_k$; \emph{GT boxes + SAM~3.1} feeds ground-truth boxes to the segmentor at the same refresh points and bounds what mask decoding alone can reach.

\begin{table*}[!t]
\centering
\caption{\textbf{Results on existing benchmarks.}  MeViS v2 results are in Appendix~\ref{app:mevis_results}.}
\label{tab:main_results}
\label{tab:existing_benchmarks}
\begingroup
\scriptsize
\setlength{\tabcolsep}{5pt}
\renewcommand{\arraystretch}{1.08}
\begin{tabular*}{\textwidth}{@{}l@{\extracolsep{\fill}}crr@{}}
\toprule
\textbf{Method / Configuration} & \textbf{Online} & \textbf{ReasonVOS} & \textbf{ReVOS (Overall)} \\
\midrule
VISA (13B, IT)~\citep{yan2024visa} & \ding{55} & -- & 50.90 \\
GLUS (7B)~\citep{lin2025glus} & \ding{55} & 49.90 & 54.90 \\
CoT-RVS (GPT-4o, offline)~\citep{kao2026cotrvs} & \ding{55} & 65.50 & 55.90 \\
Refer-Agent (9B)~\citep{jiang2026referagent} & \ding{55} & 69.80 & 61.30 \\
VideoSEG-O3 (4B)~\citep{dai2026videoseg} & \ding{55} & 62.90 & 67.70 \\
OnlineRefer~\citep{wu2023onlinerefer,bai2024videolisa} & \ding{51} & 38.70 & -- \\
CoT-RVS (LLaVA-1.5-7B, online)~\citep{kao2026cotrvs} & \ding{51} & 52.00 & 46.20 \\
ORVOS~\citep{liu2026orvos} & \ding{51} & -- & 54.30 \\
\midrule
\textbf{Sekai (ours)} & \ding{51} & 59.00 & 57.29 \\
\bottomrule
\end{tabular*}
\endgroup
\end{table*}

\begin{wrapfigure}{R}{\sidepanelwidth}
\centering\small
\includegraphics[width=0.8\linewidth,trim=0.25bp 0.5bp 0.5bp 3.25bp,clip]{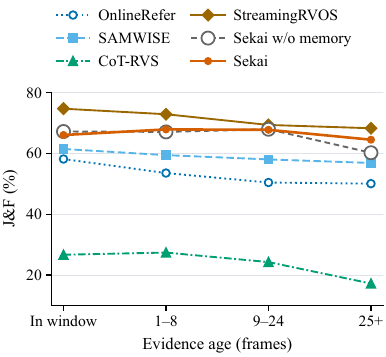}
\caption{\textbf{Performance by evidence age on Isekai.} J\&F over nonempty scoring points in the four strata, averaged within videos then across videos. All methods use the same points in each stratum.}
\label{fig:evidence_age_results}
\end{wrapfigure}
\subsection{Main Results}
\label{sec:main_results}
\paragraph{Isekai.}
Sekai obtains the best overall J\&F and the best score on \textsc{Ever} (Table~\ref{tab:isekai_online}). The remaining family columns are split among the baselines, and Table~\ref{tab:target_count} shows why: CoT-RVS is strong on empty points but collapses on single-target points, StreamingRVOS the reverse, so each leads the families whose empty-frame proportion (Figure~\ref{fig:isekai_landscape}) matches its bias toward withholding or emitting a mask. Sekai is the only method without a collapsed regime across empty, single-, and multi-target points, which is what the overall score rewards.

\paragraph{Where the gain comes from.}
The gain is concentrated on \textsc{Ever}, where Sekai leads the strongest baseline by 9.28 and the memory-free control by 4.58: an \textsc{Ever} target stays eligible after its event ends and its evidence leaves the window, which is exactly the state the EMR carries. The control also trails on \textsc{Now} and the compound families but is higher on \textsc{Latest} and \textsc{Never}, where the same memory becomes a liability: a stale or spurious entry is never revisited, so on \textsc{Latest} an older participant is not superseded and on \textsc{Never} one wrong entry excludes an instance for the rest of the video (deltas in Appendix~\ref{app:family_deltas}). On the history subset Sekai is above the control and within 3 points of StreamingRVOS. With ground-truth boxes the same segmentor reaches 87.80 (Table~\ref{tab:diagnostics}), so the remaining error lies mostly in target selection, not mask decoding.

\paragraph{Existing benchmarks.}
\label{sec:existing_benchmarks}
On ReasonVOS and ReVOS, which fix target identity per video and so test referring and segmentation quality, Sekai is the highest online entry (Table~\ref{tab:main_results}).

\subsection{Ablations and Analysis}
\label{sec:mechanism}

\paragraph{Evidence age.}
Figure~\ref{fig:evidence_age_results} scores every method on the same nonempty points per stratum (values in Table~\ref{tab:age_details}). Every baseline declines with evidence age; the memory-free control holds through 24 frames and drops 7.7 points at 25+, while Sekai is the only method whose curve is flat. The crossovers have a mechanical reading: for two or three chunks the qualifying instance is usually still the salient actor in $X_k$, so an appearance prior recovers it without any record of the event; beyond that, only the record identifies who acted. Two caveats bound this. The strata exclude empty frames and so reward emitting a mask: StreamingRVOS, which almost never withholds one, leads every stratum, and even the GT-box oracle scores 41.43 in window because a target confirmed mid-chunk is not prompted until the next endpoint, so the in-window stratum measures anticipation as much as reasoning. Conversely, Table~\ref{tab:target_count} shows the memory is neutral on nonempty points and gains 11.11 on empty ones; the strata therefore show only half of its contribution, retaining old-evidence targets, and cannot register the other half, withholding a mask when nothing visible qualifies.

\Needspace{13\baselineskip}
\begin{wraptable}{r}{\sidepanelwidth}
\begin{minipage}[t]{\sidepanelwidth}
\vspace{0pt}
\centering\small
\captionof{table}{\textbf{Module ablations on Isekai.}  }
\label{tab:diagnostics}
\scriptsize
\setlength{\tabcolsep}{1.2pt}
\begin{tabular*}{\linewidth}{@{}l@{\extracolsep{\fill}}lrr@{}}
\toprule
Module & Condition & \shortstack{All\\J\&F} & \shortstack{Hist.\\J\&F} \\
\midrule
--- & Sekai & 50.33 & 65.97 \\
EMR & w/o memory & 45.61 & 63.93 \\
EMR & No state labels & 48.57 & 62.87 \\
MGD & Unadapted locator & 27.66 & 40.34 \\
MGD & SAM2.1 segmentor & 47.92 & 60.26 \\
\midrule
\multicolumn{2}{@{}l}{GT boxes + SAM~3.1} & 87.80 & 90.85 \\
\bottomrule
\end{tabular*}
\end{minipage}

\end{wraptable}

\paragraph{Module ablations.}
Table~\ref{tab:diagnostics} ablates one module at a time. 
\emph{EMR.} Removing the memory costs 4.72 J\&F. Removing only the state labels costs 1.76 overall, since wording and evidence frames still encode most of the state, but on the history subset a memory without states falls below no memory: an entry that no longer says whether its event ended reads as still active, so a record that omits state can be worse than none.

\WFclear
\emph{MGD.} The unadapted locator loses 22.67 on the same memory, and its two scores locate what the adaptation teaches. The base model finds the instance the memory names but cannot decline to emit a box; the LoRA's first effect is abstention, the same regime where the memory contributes most. The segmentor matters least: SAM2.1~\citep{ravi2024sam2} costs 2.41, whereas backends that hold an anchor mask without propagating fall to 24--29 (Appendix~\ref{app:segmentor_swap}). With prompts only at endpoints and every frame scored, the segmentor buys coverage of the frames between endpoints, not a better endpoint mask.

\WFclear
\subsection{Qualitative Results}
\label{sec:qualitative}
\begin{figure}[!t]
\centering
\includegraphics[width=0.95\textwidth,trim=0.25bp 5bp 0.25bp 0bp,clip]{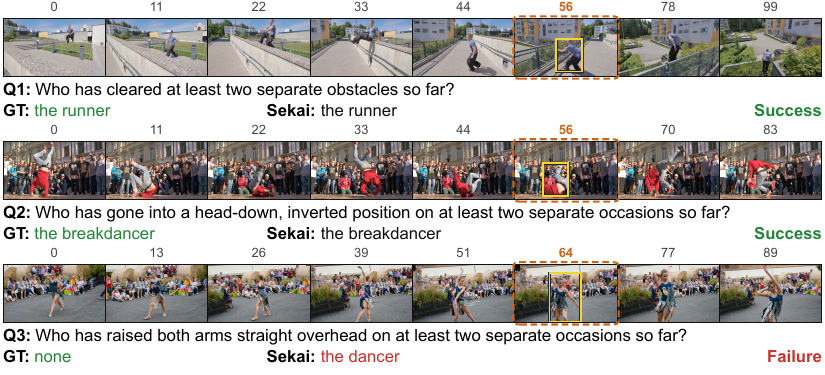}
\caption{\textbf{Qualitative examples on Isekai.} Orange dashed outlines mark the sixth-frame endpoint; yellow boxes are the MGD locator's saved outputs. }
\label{fig:case_study}
\end{figure}

\nopagebreak[4]
\paragraph{Case Study.}
Figure~\ref{fig:case_study} shows two correct selections and one failure from the run in Table~\ref{tab:isekai_online}, all \textsc{Count} queries (``at least two separate occasions''), the family that exercises both modules: the writer must keep the occurrences as separate closed entries, and the locator must count entries rather than read the chunk, since an eight-frame chunk typically shows at most one occurrence and reveals that it is the second. In the first two rows the MGD selects the runner and the breakdancer at frame 56; a chunk-only reader can make these selections only by guessing the salient actor, the behavior the 9--24 stratum rewards and the 25+ stratum exposes.
\paragraph{Failure Analysis.}
In the third row, Sekai selects the dancer at frame 64, although the label turns positive only at frame 79; the error is one of timing, not identity. Both modules err. The EMR stores the arm-raising as a single \textsc{Active} entry with evidence frames 0 and 64: because an entry closes only on observed ending evidence, a missed ending lets the writer extend one entry rather than open a second, merging the ``separate occasions'' the query asks about. \textsc{Count} on this memory gives one, so a faithful readout would have withheld a box; that the MGD emitted one indicates it fell back to the chunk, where the raised arms are visible. Since the locator does not emit the IDs it selects, the two errors cannot be separated in the output; the case instantiates the pattern behind the \textsc{Latest} and \textsc{Never} deficits in Table~\ref{tab:isekai_online}, a record wrong once and never revisited, and motivates both the ID readout and the re-verification step of Section~\ref{sec:limitations}. %

\section{Limitations and Conclusion}
\label{sec:limitations}
\paragraph{Limitations.}
Sekai's carried state is a text memory written by a frozen model, and the design accepts a trade-off: what the writer records stays available indefinitely, but what it misrecords is not revisited, since later calls cannot re-examine frames that have left the window. Knowledge accumulated in increments also accumulates its errors. Table~\ref{tab:isekai_online} shows both sides: the same memory that carries \textsc{Ever} targets past their evidence can also hold a stale entry that hurts \textsc{Latest} and \textsc{Never}, so a verification step that re-checks entries against new evidence is the natural next component. The locator, adapted partly on synthetic memories, selects from the memory without emitting the IDs it selects; an explicit ID readout would make cases such as the third row of Figure~\ref{fig:case_study} diagnosable at the module level. 

\paragraph{Conclusion.}
GRVS treats the streaming target set as a function of the observed event history that may be empty, single, or plural at any frame. Isekai measures this frame by frame with event-derived targets and evidence-age strata. Sekai shows that a bounded text record of who did what, and when, retains targets after their evidence leaves the window, with its clearest margin over prior methods on EVER queries and, relative to a memory-free control, on the oldest evidence. Keeping that record correct, and reading it for recency and negation, remain open.
\FloatBarrier
\subsection*{Generative AI Usage Statement}
We used generative AI tools to help organize and edit the manuscript, search the literature, and write plotting code. The authors designed and ran all experiments and checked all technical content and results.

\subsection*{Ethics Statement}
The work uses existing video benchmarks and their annotations, with data sources and usage conditions recorded alongside the assets. Models that bind past events to visible people or objects can make consequential identity errors; deployment in sensitive settings requires evaluation beyond the present research benchmarks.

\subsection*{Reproducibility Statement}
Appendix~\ref{app:implementation} summarizes the prompts (full text released with the code), locator training data and hyperparameters; Appendix~\ref{app:baselines} describes the causal inputs for every baseline. The code, including the full prompt text, evaluation scripts, and example Isekai annotations, is available at the public repository linked in the abstract. The full Isekai annotations will be released upon publication.

\bibliography{iclr2027_conference}
\bibliographystyle{iclr2027_conference}
\clearpage
\appendix
\section{Isekai Construction and Review}
\label{app:isekai}

This appendix gives the release details, review safeguards, and scoring conventions for Isekai. The main text reports the 60 videos, 197 fixed queries, and 12,569 scored query--frame records; this appendix records the construction and scoring details.

\subsection{Per-family definitions and safeguards}
\label{app:isekai_definitions}
\begin{table}[!t]
\centering\small
\caption{\textbf{Isekai query families.} Primary families apply a single predicate to the reviewed event history; compound families compose temporal conditions with counting, ordering, negation, or duration. $n$: number of fixed queries; scoring points count query--frame pairs marked for scoring.}
\label{tab:families}
\setlength{\tabcolsep}{4pt}
\begin{tabularx}{\linewidth}{@{}>{\raggedright\arraybackslash}p{0.20\linewidth}>{\raggedright\arraybackslash}p{0.19\linewidth}rr>{\raggedright\arraybackslash}X@{}}
\toprule
Family & Meaning & $n$ & \shortstack{Scoring\\points} & Example \\
\midrule
\textsc{Now} & Participates now & 60 & 3,821 & ``Which dog is rearing up?'' \\
\textsc{Ever} & Participated so far & 60 & 3,820 & ``Which dogs have reared up so far?'' \\
\textsc{Latest} & Participant of the most recently begun occurrence & 18 & 1,151 & ``Which dog most recently reared up?'' \\
\textsc{Never} & No occurrence confirmed so far & 21 & 1,339 & ``Which dog has not reared up?'' \\
\midrule
\textsc{First} & Participant of the earliest-begun occurrence & 5 & 306 & ``Which dog reared up first?'' \\
\textsc{Past-not-present} & Participated before, but not now & 4 & 246 & ``Which dogs have reared up but are not rearing now?'' \\
\textsc{Count} & Number of occurrences & 9 & 676 & ``Which dog reared up at least twice?'' \\
\textsc{Two-hop} / \textsc{Object-history} & Participant roles across events & 7 / 6 & 460 / 372 & ``first player to shoot after dribbling'' \\
\textsc{Before/After} / \textsc{Continuous} & Event order / duration & 4 / 3 & 186 / 192 & ``Who reared up before the other dog?'' / ``Who kept moving?'' \\
\bottomrule
\end{tabularx}
\end{table}

\textsc{Now} selects currently participating instances. \textsc{Ever} retains every participant confirmed so far. \textsc{Latest} selects the participant of the most recently begun occurrence, and \textsc{First} selects the participant of the earliest-begun occurrence. \textsc{Past-not-present} combines a past event with a known current nonparticipation state. \textsc{Count}, \textsc{Object-history}, \textsc{Two-hop}, \textsc{Before/After}, and \textsc{Continuous} add counting, participant roles, relation chains, ordering, or interval constraints. \textsc{Never} is the prefix complement of \textsc{Ever}, $s^{\textsc{Never}}_{i,t}=1-s^{\textsc{Ever}}_{i,t}$, so its set contracts as qualifying events are confirmed.

Using the event notation in Section~\ref{sec:problem_setup}, the temporal predicates are
\begin{equation}
\begin{gathered}
s^{\textsc{Now}}_{i,t}=\mathbf{1}\!\left\{\exists e\in\mathcal{E}_t:\ i\in P_e,\ a_e\le t\le b_e\right\},\qquad
s^{\textsc{Ever}}_{i,t}=\mathbf{1}\!\left\{\exists e\in\mathcal{E}_t:\ i\in P_e\right\},\\
s^{\textsc{Latest}}_{i,t}=\mathbf{1}\!\left\{i\in P_{e^\star_t}\right\},\quad e^\star_t=\arg\max\nolimits_{e\in\mathcal{E}_t} a_e,\\
s^{\textsc{First}}_{i,t}=\mathbf{1}\!\left\{i\in P_{\underline e_t}\right\},\quad \underline e_t=\arg\min\nolimits_{e\in\mathcal{E}_t} a_e,\qquad
s^{\textsc{Never}}_{i,t}=1-s^{\textsc{Ever}}_{i,t}.
\end{gathered}
\label{eq:semantics}
\end{equation}
Crucially, while typically $\tau_e \in [a_e, b_e]$, fast transitional events (e.g., a brief jump confirmed only upon landing) can have $\tau_e > b_e$; in strict causal streaming, such an event never enters \textsc{Now} but commits directly to \textsc{Ever} and \textsc{Latest} once confirmed at $\tau_e$. When $\mathcal{E}_t$ is empty, \textsc{Latest} and \textsc{First} are false for every instance. The remaining families (Section~\ref{sec:query_families}) compose these predicates with counting, ordering, or duration. $\mathcal{E}_t$ changes only at confirmation frames $\tau_e$, and \textsc{Now} additionally at ending frames $b_e+1$, so $\mathcal{A}_t$ changes only at event boundaries.

For the compound families, \textsc{Past-not-present} is $s^{\textsc{Ever}}_{i,t}\wedge\neg s^{\textsc{Now}}_{i,t}$, and \textsc{Count} tests $|\{e\in\mathcal{E}_t:i\in P_e\}|\ \{=,\ge\}\ c$. \textsc{Two-hop} and \textsc{Object-history} follow a role chain across events; \textsc{Before/After} and \textsc{Continuous} impose order and interval constraints, respectively.

We require prefix-confirmed evidence for every positive event and observed evidence of ending before marking an event ended. Disappearance alone cannot certify an ending. We distinguish known absence, qualified-but-invisible instances, insufficient evidence, and missing annotation. The last two are withheld from determinate scoring.

\paragraph{Counterparts on the EMR memory.}
The predicates have direct counterparts on $M_k$ (Section~\ref{sec:system_record}): \textsc{Now} checks for an \textsc{active} entry, \textsc{Ever} for any entry, \textsc{Never} for none, \textsc{Latest} and \textsc{First} pick the owner of the entry with the latest or earliest first evidence, and \textsc{Count} counts entries.

\subsection{Review procedure and release gate}
\label{app:isekai_review}
Machine proposals provide candidate object--condition intervals. Reviewers inspect continuous frames, identify participants, correct inclusive boundaries, and inspect every object's complete video, including objects that never qualify. They record participant roles, first prefix-confirmable evidence, endings, evidence frames, and uncertainty. Reviewers may inspect the full video to locate an event, but each label must be justified from the observed prefix. Concurrent or unresolved order remains marked unresolved.

For each scored $(q,t)$, we derive \texttt{eligible\_ids} from reviewed events and \texttt{visible\_ids} from official masks, then store their intersection as \texttt{output\_ids}. Reference IDs link event participants to masks. The release gate checks answer transitions, empty and multiple targets, ordinary sampled episodes, source hashes, and mask availability.

\subsection{Inter-annotator agreement}
\label{app:isekai_agreement}
A blind independent audit compares annotations on 20 sampled videos. Table~\ref{tab:annotation_agreement} reports the event-level comparison and the recompiled scoring-point comparison. The scoring-point audit covers \textsc{Now}, \textsc{Ever}, and \textsc{Never} only. Latest and compound query families are omitted because the independent export does not separately record the required answer and evidence states or query-specific intervals.
\begin{table}[!t]
\centering
\small
\caption{\textbf{Independent annotation audit.} Event matching uses same video and event type, overlapping intervals, and exact participant--role sets. Participant agreement uses the relaxed same-type, overlapping-interval pairing.}
\label{tab:annotation_agreement}
\begin{tabular}{p{0.70\linewidth}r}
\toprule
Measure & Value \\
\midrule
\multicolumn{2}{l}{\textit{Event annotations (20 audited videos; 36 independent, 34 prior events)}} \\
Strictly matched events & 29 \\
Independent-event match rate & 80.6\% \\
Prior-event coverage & 85.3\% \\
Mean interval IoU among matched events & 0.916 \\
Matched intervals with IoU $\geq 0.5$ & 100\% \\
Matched pairs crossing a chunk endpoint boundary & 9 \\

\midrule
\multicolumn{2}{l}{\textit{Primary causal labels (2,307 scoring points)}} \\
Exact output-set agreement & 93.0\% \\
Mean output-set Jaccard & 0.940 \\
Empty/nonempty accuracy & 94.9\% \\
Empty/nonempty Cohen's $\kappa$ & 0.962 \\
\bottomrule
\end{tabular}
\end{table}

\subsection{Scoring details}
\label{app:isekai_exceptions}
All main results use the 12,569 scoring points. At each point, reference masks for the selected visible instances are unioned, as are predicted masks. Region IoU and boundary F are evaluated on these binary foregrounds. Two empty unions score one and a one-sided empty union scores zero; invalid or missing predictions count as empty. We average within each video and then equally over videos. This score does not assess instance separation. 

\subsection{Evidence age}
\label{app:isekai_age}
For a scoring frame $t$, let $c(t)=0$ at initialization and $c(t)=1+8\lfloor(t-1)/8\rfloor$ otherwise. The current observed chunk is $x_{c(t):t}$. Restrict the reviewed event history to $\mathcal E_t^{\mathrm{win}}=\{e\in\mathcal E_t:\tau_e\ge c(t)\}$. A point belongs to the replayable history subset when
\[
\{i\in\mathcal A_t(\mathcal E_t):v_{i,t}=1\}
\ne
\{i\in\mathcal A_t(\mathcal E_t^{\mathrm{win}}):v_{i,t}=1\}.
\]
The full-history replay agrees with all 8,980 labels for \textsc{Now}, \textsc{Ever}, and \textsc{Never}. For an ongoing \textsc{Now} event, its first confirmation can precede the chunk even when the action remains visible; the history criterion concerns event replay rather than visual impossibility.

For a supporting event $e$, evidence age is $t-\tau_e$. An event confirmed in the current chunk is assigned to \emph{in window}; remaining points fall into 1--8, 9--24, or 25+ frames. \textsc{First} uses the earliest qualifying event and \textsc{Latest} the most recent. Repeated \textsc{Ever} events use the latest qualifying occurrence for each participant. A multi-target point uses the greatest participant age and is counted once.

\paragraph{Dataset-landscape timing.}
\label{app:isekai_landscape_timing}
Figures~\ref{fig:isekai_landscape} and~\ref{fig:evidence_age_results} use the same timing assignments. Temporal order combines \textsc{Latest}, \textsc{First}, and \textsc{Before/After}; Other history combines \textsc{Count}, \textsc{Two-hop}, \textsc{Object-history}, \textsc{Past-not-present}, and \textsc{Continuous}. Target counts use distinct visible reference IDs. The bars and performance curves contain 7,190 nonempty points, including 1,028 multi-target points; the landscape proportions use all 12,569 points. Event-age strata group nonempty labels by evidence timing, whereas the history subset includes empty references whenever window-only event replay changes the target set.

Where an explicit supporting-event timing is unavailable, the current answer interval's onset supplies a timing proxy (1,621 points). For \textsc{Never}, the anchor is the earliest confirmed exclusion event (918 points) or the video start when no such event exists (400 points). Temporary invisibility does not reset an answer interval. The four strata contain 972, 1,211, 2,522, and 2,485 points; Appendix~\ref{app:evidence_age_records} reports their scores.

\subsection{Teaser provenance}
\label{app:teaser_record}
The teaser uses DAVIS 2017~\citep{ponttuset2017davis} \emph{dogs-jump} and official visible masks for gray dog $G$ and black dog $B$. 

\section{Sekai Implementation}
\label{app:implementation}

This appendix specifies the reference-system implementation and response contracts. It records the active locator adaptation and inference procedure.

\subsection{Schedule and Module Interface}
\label{app:response_contract}
The EMR's writer and the MGD's locator use Qwen3-VL-8B-Instruct~\citep{bai2025qwen3vl}; SAM~3.1~\citep{carion2025sam3} is frozen. Calls follow the first-frame schedule in Section~\ref{sec:method}: initialize on $x_0$, then process $X_k=x_{a_k:b_k}$ with $a_1=b_1=0$ and $a_k=b_{k-1}+1$. A final partial chunk is processed if the video ends off-schedule. The writer reads $q$, $X_k$, and $M_{k-1}$ and replaces the EMR memory. The locator reads $q$, $X_k$, and the new memory in a fresh context. Both the EMR memory and locator rendering are validated against the 128-token budget. Event indices record first and most recent evidence, not guaranteed event boundaries.

\subsection{Event Memory Reasoner}
\label{app:emr}
\label{app:prompts}
The active writer instruction requires stable local IDs, observed appearance and locations, event labels, event states, evidence indices, complete replacement output, and a 128-token limit. It forbids inventing events or treating disappearance as an ending.

An over-budget output is rejected and regenerated, with at most three identical attempts. Both the memory and its locator rendering must fit within 128 tokens. In the 1,770 writer-call logs, 11 calls (0.7\%) encountered the budget limit. After repeated failure, the previous accepted memory and segmentation state are carried forward; the failure remains in the evaluation denominator.

\subsection{Memory-Grounded Decoder}
\label{app:locator_config}
Grammar-constrained decoding validates the single top-level JSON response. The locator always returns a list of boxes, including an empty list. Coordinates use normalized $[0,1000]$ xyxy values, and malformed, truncated, or degenerate boxes are invalid.

The active reference run freezes the writer, vision backbone, token embeddings, and segmentor, and adapts the locator with response-only cross-entropy. It uses 3,579 real samples and 2,592 synthetic samples from the released MeViS v2 training corpus. Real samples pair queries and current chunks with writer-generated memory states and annotated endpoint boxes; synthetic samples pair object descriptions and endpoint geometry with hypothetical temporal states. The main locator uses LoRA rank 16, scaling 32, dropout 0.05, learning rate $2\times10^{-5}$, and effective batch size 4 (two real and two synthetic samples).

The active locator instruction asks for every visible instance satisfying the original query in the last frame and permits an empty box list. Full prompt text, validators, and exact serialization are included in the public repository.

\paragraph{Video-level separation.}
We manually checked the complete source-video manifests for Isekai and locator training, and reviewed all candidate overlaps identified by the available-frame duplicate checks. To the best of our knowledge, none of the 60 Isekai videos overlaps with the source videos used for locator training, including the sources of real samples and the video geometry used in synthetic samples. This check concerns locator-training data; it does not establish separation from foundation-model pretraining data.

\subsection{Inference Procedure}
\label{app:inference_procedure}
\begin{enumerate}
\setlength{\itemsep}{0pt}
\setlength{\parsep}{0pt}
\item Initialize empty semantic memory and segmentation state.
\item At each reasoning endpoint, run the EMR (writer) on the arrived chunk, the query, and the previous memory; validate the replacement memory.
\item In a fresh context, run the MGD's locator on the new memory and the same chunk; validate its endpoint boxes.
\item Use the MGD's segmentor to apply valid boxes to the endpoint frame. A valid empty list clears active segmentation state while preserving semantic memory.
\item Use the MGD's segmentor to propagate accepted masks on intervening arrived frames and carry the memory to the next endpoint.
\end{enumerate}

\subsection{Segmentor replacement and GT-box oracle}
\label{app:segmentor_swap}
\begin{table}[!t]
\centering\small
\caption{\textbf{Segmentor replacement on Isekai.} Video-equal union-mask scores (\%) on all released scoring points. The four replacement backends share locator boxes. Hold copies the anchor mask until the next refresh; Propagate updates masks on arriving frames. The GT-box row supplies privileged prompts; its family and history-subset J\&F scores are shown below.}
\label{tab:segmentor_swap}
\begin{tabular}{@{}llrrr@{}}
\toprule
Backend & Between anchors & $\mathcal J$ & $\mathcal F$ & $\mathcal J\&\mathcal F$ \\
\midrule
Filled rectangle & Hold & 28.39 & 20.57 & 24.48 \\
GrabCut & Hold & 30.00 & 27.35 & 28.67 \\
SEEM (FocalT-v0) & Hold & 29.73 & 28.78 & 29.26 \\
SAM2.1 (Hiera-Large) & Propagate & 46.47 & 49.36 & 47.92 \\
GT boxes + SAM 3.1 & Propagate & 87.28 & 88.31 & 87.80 \\
\bottomrule
\end{tabular}
\par\medskip
\begin{tabular}{@{}lrrrrrr@{}}
\toprule
Prompts & Now & Ever & Latest & Never & Other & Hist. \\
\midrule
GT boxes & 84.53 & 89.27 & 82.64 & 91.53 & 91.12 & 90.85 \\
\bottomrule
\end{tabular}
\end{table}

The four locator-prompted branches reuse saved outputs with the saved writer memories.

As in the main evaluation, the scorer unions instance masks within each frame, computes region IoU and boundary F, averages scored frames within each video, and then gives all 60 videos equal weight. Two empty masks score one; a one-sided empty match scores zero. Family scores use the same video-equal aggregation after filtering; Other pools the compound families, excluding Now/Ever/Latest/Never before aggregation. Union-mask scoring does not measure instance separation.

Filled rectangles, GrabCut, and SEEM generate a mask at each locator anchor and hold it until the next anchor. GrabCut falls back to a filled rectangle on an OpenCV error; the receipts do not count these internal fallbacks separately. SEEM uses the official FocalT-v0 weights and \texttt{transformer\_encoder\_fpn} configuration. Each locator box is rasterized into the official spatial-stroke prompt interface. SAM2.1 Hiera-Large instead propagates within arrived segments; only the audited causal implementation is included. Thus the comparison changes propagation as well as the anchor-frame segmentor.

The SAM~3.1 oracle receives ground-truth boxes at the same causal refresh schedule and propagates through arriving frames; it does not read ground-truth masks during inference. Its 87.80 score measures performance with privileged target prompts.
\section{Baseline Adaptations and Causal Audit}
\label{app:baselines}

We adapt each baseline to the same arrived-frame schedule before comparing it on Isekai. The audit records the frames, features, caches, and prompts available at every call and rejects any future-derived input.

\subsection{OnlineRefer and CoT-RVS}
\label{app:baseline_reasoning}
OnlineRefer~\citep{wu2023onlinerefer} receives the fixed query and newly arrived frames while retaining its target representation. CoT-RVS~\citep{kao2026cotrvs} receives a causal prefix grid at each refresh and uses its native reasoning and segmentation interface. The online branch follows a single object of interest and can switch targets. OnlineRefer uses the official Swin-L checkpoint trained on Refer-YouTube-VOS; our CoT-RVS adaptation uses Qwen3-VL-8B-Instruct and SAM~3.1.

\subsection{StreamingRVOS and SAMWISE}
\label{app:baseline_streaming}
StreamingRVOS~\citep{zhang2026streamingrvos} carries semantic tokens between refreshes and is evaluated with its causal streaming schedule. SAMWISE~\citep{cuttano2025samwise} receives only its recent visual context and segmentation memory. We keep their native mask decoders. StreamingRVOS uses the official 4B checkpoint and SAMWISE the official MeViS HSA+CME checkpoint; both include MeViS training.

\subsection{Controls}
\label{app:baseline_controls}
Sekai w/o memory removes the EMR, so the locator receives only the query and current chunk. GT boxes + SAM~3.1 supplies reference boxes at the endpoint and measures mask generation after target selection.

\subsection{Comparison with a Shared Backbone and Segmentor}
\label{app:shared_backbone}

Our causal adaptation of CoT-RVS~\citep{kao2026cotrvs} and Sekai
both use Qwen3-VL-8B-Instruct~\citep{bai2025qwen3vl} and
SAM~3.1~\citep{carion2025sam3}. They follow the same arrival schedule
and are evaluated on the same 12,569 scoring points.
CoT-RVS reads a causal-prefix grid and supplies text prompts to the
segmentor. Sekai reads incremental memory with an adapted locator
and supplies box prompts. This comparison holds the backbone family
and segmentor fixed; locator adaptation and prompting differ.

\begin{table}[!t]
\centering
\small
\caption{\textbf{Shared-backbone and shared-segmentor comparison.}
Scores on Isekai (\%).}
\label{tab:shared_backbone}
\begin{tabular}{@{}lrrr@{}}
\toprule
Method & J & F & J\&F \\
\midrule
CoT-RVS, causal adaptation & 46.43 & 47.90 & 47.16 \\
Sekai & \textbf{49.09} & \textbf{51.58} & \textbf{50.33} \\
\bottomrule
\end{tabular}
\end{table}
\section{Supplementary Diagnostic Records}
\label{app:diagnostics}

\subsection{Evidence-age summaries}
\label{app:evidence_age_records}
Table~\ref{tab:age_details} gives the values plotted in Figure~\ref{fig:evidence_age_results} and the GT-box reference. Each stratum uses the same nonempty reference points for every method. Scores are averaged within videos and then across videos represented in that stratum.
\begin{table}[!t]
\centering\small
\caption{J\&F by evidence age. Counts are scoring points.}
\label{tab:age_details}
\begin{tabular}{@{}lrrrr@{}}
\toprule
Method & In window & 1--8 & 9--24 & 25+ \\
\midrule
Points & 972 & 1,211 & 2,522 & 2,485 \\
\midrule
OnlineRefer & 58.08 & 53.47 & 50.37 & 49.99 \\
SAMWISE & 61.38 & 59.35 & 57.95 & 56.78 \\
CoT-RVS & 26.64 & 27.36 & 24.23 & 17.19 \\
StreamingRVOS & 74.61 & 72.79 & 69.27 & 68.19 \\
Sekai w/o memory & 67.11 & 66.99 & 67.86 & 60.14 \\
Sekai & 65.96 & 67.81 & 67.67 & 64.42 \\
GT boxes + SAM 3.1 & 41.43 & 91.88 & 90.26 & 92.51 \\
\bottomrule
\end{tabular}
\end{table}

\subsection{Target-count summaries}
\label{app:target_count}
Table~\ref{tab:target_count} groups scoring points by the number of distinct visible reference instances. We filter to each group, average scores within each video, and then average over videos represented in that group, using the same saved predictions as Table~\ref{tab:isekai_online}.
\begin{table}[!t]
\centering\small
\caption{\textbf{J\&F by target count on Isekai.} Counts refer to visible reference instances. Best per column in bold, excluding the privileged GT-box reference.}
\label{tab:target_count}
\begin{tabular}{@{}lrrr@{}}
\toprule
Method & $0$ targets & $1$ target & $\geq2$ targets \\
\midrule
Points & 5,379 & 6,162 & 1,028 \\
\midrule
OnlineRefer~\citep{wu2023onlinerefer} & 4.60 & 53.32 & 34.82 \\
SAMWISE~\citep{cuttano2025samwise} & 4.92 & 59.13 & 44.66 \\
CoT-RVS~\citep{kao2026cotrvs} & \textbf{75.76} & 21.45 & 30.07 \\
StreamingRVOS~\citep{zhang2026streamingrvos} & 5.63 & \textbf{70.74} & \textbf{50.91} \\
\midrule
Sekai w/o memory & 17.14 & 65.39 & 49.44 \\
Sekai & 28.25 & 65.22 & 49.60 \\
\midrule
GT boxes + SAM 3.1 & 93.38 & 80.17 & 89.11 \\
\bottomrule
\end{tabular}
\end{table}

\subsection{Per-Family Comparison with the Memory-Free Control}
\label{app:family_deltas}
Table~\ref{tab:isekai_online} gives the absolute scores; here we list the differences that Section~\ref{sec:main_results} summarizes. Relative to Sekai w/o memory, Sekai changes J\&F by $+7.44$ on \textsc{Now}, $+4.58$ on \textsc{Ever}, $-5.97$ on \textsc{Latest}, $-10.51$ on \textsc{Never}, $+11.99$ on the pooled OTHER, and $+2.04$ on the history subset. Relative to the strongest baseline in each column, Sekai is $+9.28$ on \textsc{Ever} (StreamingRVOS), $-17.27$ on \textsc{Now} and $-5.21$ on Other (CoT-RVS), and $-2.50$ on \textsc{Latest}, $-11.17$ on \textsc{Never}, and $-2.74$ on the history subset (StreamingRVOS). The two families where the memory hurts share one failure mode: a stale or spurious entry that cannot be corrected once its frames have left the window. \textsc{Latest} requires the newest event to supersede an older participant, and \textsc{Never} excludes an instance permanently once any entry records the event, so a single false positive in the memory costs every later frame.

\paragraph{Uncertainty of the overall comparison.}
We estimate uncertainty by paired bootstrap over the 60 videos. Each replicate samples 60 videos with replacement and uses the
same sampled videos for both methods, retaining all queries and scoring points within each video. With 100,000 replicates
, Sekai improves over the adapted CoT-RVS baseline by 3.163 J\&F points, with a 95\% percentile confidence interval of $[0.05,\,6.27]$. 

\subsection{History-subset coverage}
\label{app:controlled_records}
The 4,133 replayable history points comprise 866 \textsc{Now}, 2,401 \textsc{Ever}, and 866 \textsc{Never} points. We first replay each full-history query and check equality with its visible reference set, then retain only events first confirmed within the current chunk and compare the two sets. Appendix~\ref{app:isekai_age} defines the criterion. No \textsc{Latest} or compound point is assigned to this subset without a replayable query rule.

\subsection{MeViS v2 results}
\label{app:mevis_results}
Table~\ref{tab:mevis_results} reports the MeViS v2 results. Sekai uses the official complete validation-set evaluation and the locator configuration described in Appendix~\ref{app:locator_config}.
\begin{table}[!t]
\centering\small
\caption{\textbf{Results on MeViS v2.} J\&F (\%). Reference results are as reported by \citet{ding2025mevis}.}
\label{tab:mevis_results}
\begin{tabular}{@{}lcr@{}}
\toprule
Method / Configuration & Online & J\&F \\
\midrule
LMPM~\citep{ding2023mevis} & \ding{55} & 38.30 \\
DsHmp~\citep{he2024dshmp} & \ding{55} & 40.80 \\
LMPM++~\citep{ding2025mevis} & \ding{55} & 43.90 \\
VISA~\citep{yan2024visa} (MeViS v2 evaluation) & \ding{55} & 40.70 \\
\textbf{Sekai (ours)} & \ding{51} & 50.45 \\
\bottomrule
\end{tabular}
\end{table}

\end{document}